# Weakly Supervised Learning with Automated Labels from Radiology Reports for Glioma Change Detection

Tommaso Di Noto[1,*], Meritxell Bach Cuadra[1,2], Chirine Atat[1], Eduardo Gamito Teiga[1], Monika Hegi[3,4,6], Andreas Hottinger[4,5,6], Patric Hagmann[1,6], Jonas Richiardi[1]

1. Department of Radiology, Lausanne University Hospital (CHUV), and University of Lausanne (Unil)
2. CIBM, Center for Biomedical Imaging, Lausanne
3. Neuroscience Research Center, Lausanne University Hospital (CHUV) and University of Lausanne (Unil)
4. Department of Neurosurgery, Lausanne University Hospital (CHUV) and University of Lausanne (Unil)
5. Department of Oncology, Lausanne University Hospital (CHUV) and University of Lausanne (Unil)
6. Lundin Brain Tumor Research Center, Lausanne University Hospital (CHUV) and University of Lausanne (Unil)

   * corresponding author (tommydino@hotmail.it), Centre de Recherche en Radiologie PET03/02/210
   CHUV, Rue du Bugnon 46, CH-1011 Lausanne, Switzerland

*Abstract* - Gliomas are the most frequent primary brain tumors in adults. Glioma change detection aims at finding the relevant parts of the image that change over time in order to detect progression/response, and tailor treatment. Although Deep Learning (DL) shows promising performances in similar change detection tasks, the creation of large annotated datasets represents a major bottleneck for supervised DL applications in radiology. To overcome this, we propose a combined use of weak labels (imprecise, but fast-to-create annotations) and Transfer Learning (TL). Specifically, we explore inductive TL, where source and target domains are identical, but tasks are different due to a label shift: our target labels are created manually by three radiologists, whereas our source weak labels are generated automatically from radiology reports with a Natural Language Processing pipeline. We frame knowledge transfer as hyperparameter optimization, thus avoiding heuristic choices that are frequent in related works. We investigate the relationship between model size and TL, comparing a low-capacity VGG with a higher-capacity ResNeXt model. We evaluate our models on 1693 T2-weighted magnetic resonance imaging difference maps created from 183 patients, by classifying them into stable or unstable according to tumor evolution. The weak labels extracted from radiology reports allowed us to increase dataset size more than 3-fold, and significantly improve VGG classification results from 75% to 82% Area Under the ROC Curve (AUC) (p=0.04). Mixed training from scratch led to higher performance than fine-tuning or feature extraction. To assess generalizability, we also ran inference on an open dataset (BraTS-2015: 15 patients, 51 difference maps), reaching up to 76% AUC. Overall, results suggest that medical imaging problems may benefit from smaller models and different TL strategies with respect to large computer vision datasets, and that report-generated weak labels are effective in improving model performances. Code, in-house dataset and new BraTS labels are released.

# 1. Introduction

Change detection aims at spotting the parts of an image that change over time. In recent years, this task has attracted increasing attention for medical imaging applications such as multiple sclerosis [1], chest x-ray [2], retinal fundus images [3] and glioma [4]–[6], as well as in other areas such as remote sensing and video processing [7]. Change detection is particularly relevant for evolving diseases monitored longitudinally, such as gliomas, the most frequent primary brain tumors occurring in the adult population. Their most aggressive form (high-grade glioma) has a low survival (median ≤ 2 years) and requires prompt and dedicated treatment [8]. Magnetic Resonance Imaging (MRI) is the gold standard modality to monitor the evolution of gliomas since it allows the acquisition of diverse sequences, which provide complementary information to clinicians [9], and potentially avoids the need for serial biopsies [10]. Glioma change detection is a clinically-relevant, meticulous and non-trivial task for radiologists whose goal is to visually detect relevant tumor-related changes in order to detect early progressions or responses, and tailor treatment. Throughout the rest of the paper, we denote the longitudinal monitoring of gliomas via MRI as "glioma change detection".

In this work, we address glioma change detection via a Deep Learning (DL) pipeline that leverages weak labels and transfer learning. In the following paragraphs, we explain why these two expedients are useful in medical imaging, and how they can be exploited to increase performances.

The need for large amounts of manual annotations is arguably the major bottleneck for the development of supervised DL models in medical imaging. Not only is the creation of manual labels tedious for medical experts, but it is also extremely time-consuming [11]; combined with the increasing workload of radiologists [12], the creation of manual labels is expected to become more and more expensive in the coming years. This has prompted interest and advances in more sample-efficient learning methods. In this respect, weak labels are an interesting alternative to manual labels: they correspond to noisy, limited, or imprecise labels that are adopted to guide the learning process [13]–[16]. One under-explored approach for generating weak labels in medical imaging is the use of radiology reports. Most of the time, reports are stored as unstructured free-text and exhibit a strong degree of ambiguity and lack of conciseness [17]. However, recent advances in Natural Language Processing (NLP) enable the extraction of clinically-relevant labels [18]–[29] from radiology reports. Although these weak labels are inherently imperfect, they are drastically faster to obtain with respect to manual labels, and are also more scalable since they potentially allow to leverage tens of thousands of retrospective exams that would otherwise remain unused in hospital PACS (Picture Archiving and Communication Systems).

Transfer Learning (TL) is the branch of machine learning where knowledge acquired from a specific task or domain (source) is exploited to solve a downstream, related task (target) [30]. Since datasets used by most research groups in medical imaging are typically small [31] (especially compared to datasets in Computer Vision (CV)), TL holds great potential to overcome data scarcity in the field. Adopting the notation from [32], we can formally define a domain $D$ and a task $T$ as $D = \{X, P(x)\}$ and $T = \{Y, f(\cdot)\}$, where $X$ is the feature space, $P(x)$ is the corresponding marginal probability distribution, $Y$ is the label space, and $f(\cdot)$ is the objective predictive function. Moreover, we use the notations $D_s$, $D_t$, $T_s$, and $T_t$ to indicate source domain, target domain, source task and target task, respectively. Most papers dealing with medical TL focused on the choice of the source domain $D_s$, trying to understand which is the best $D_s$ from which we should transfer knowledge. For instance, several works investigated the use of natural images (e.g. the ImageNet dataset [33]) for pre-training [34]–[37]. Conversely, more recent works showed that the use of natural images could lead to negligible performance improvements [38], and rather suggested that using a medical domain as source is preferable [39]–[41]. Differently from these previous studies, in our work we explore the TL scenario where source and target domain are identical ($D_s = D_t$), but the tasks are different ($T_s \neq T_t$) because of distinct label spaces ($Y_s \neq Y_t$). In other words, we aim to understand to what extent it is possible to transfer knowledge from a source domain which has a different label distribution from the target domain, a scenario called *inductive* TL [32]. We address the task of glioma change detection with difference maps as input samples ($D_s = D_t$, details in section 2.4), with $Y_s$ consisting of weak labels generated automatically from radiology reports, and $Y_t$ consisting of manual labels created by human experts, again from radiology reports.

Once domains ($D_s$, $D_t$) and tasks ($T_s$, $T_t$) have been defined, TL can be further subdivided into three main types [42]:

- **Fine-tuning**: the DL model is pre-trained on the source domain and then all its weights are fine-tuned on the target domain.
- **Feature Extraction**: the DL model is pre-trained on the source domain and then only some of its

weights (typically the last linear layers) are fine-tuned on the target domain. Instead, the convolutional backbone layers are usually "frozen" (i.e. not trained again).
- **Mixed Training**[1]: the DL model is trained only once on a mixed dataset composed of source domain and the training portion of the target domain.

Similarly to the discussion about the choice of the source domain, there is also a lack of consensus regarding which type of TL is the most effective (e.g. is fine-tuning better than feature extraction?), with most of the works trying several combinations empirically [42]. In this paper, we develop an automated pipeline which treats the TL type as just another hyperparameter to optimize. Because the optimal value of other hyperparameters (such as the learning rate) depends on the TL type, our approach avoids the arbitrary choice of a TL type which can be potentially suboptimal.

Beside the choice of the source domain and the TL type to adopt, it is also unclear how much model size influences classification results during TL. For instance, [38] found that large networks that yield state-of-the-art results for ImageNet are not necessarily the top performing networks for medical datasets. Moreover, the authors in [38] also showed that in the small data regime (i.e. few thousands of samples or below) large ImageNet models benefit more from TL with respect to smaller networks. This behavior is frequent in CV where larger networks tend to maintain a performance edge over smaller networks even in the low-data regime [43]. Subsequent work [44] instead found that using extremely large architectures (380M parameters) and massive pre-training datasets (300M images) from the natural images domain can actually improve results on target medical domains. In the same line as these studies, we compare a low-capacity model to a higher-capacity model (details in section 2.5) to investigate the impact of model size, but in the scenario of inductive TL.

In summary, the goal of this paper is to tackle glioma change detection within an inductive TL scenario ($D_s = D_t$, $Y_s \neq Y_t$). The main contributions of our work are the following: (i) we propose a TL approach that leverages inexpensive and fast-to-create weak labels generated from radiology reports; (ii) we automate the choice of TL type, treating it as another hyperparameter to optimize, and thus avoiding manual empirical trials; (iii) we assess the impact of model size on TL for medical imaging and (iv) we release new expert labels for the longitudinal subjects of the public BraTS 2015 dataset [45]–[47], as well as our in-house 1693 longitudinal difference images for glioma, the largest such dataset currently available.

## 1.1 Related works

Previous works have addressed the task of glioma change detection. For instance, [4] used difference maps after contrast midway mapping to monitor tumor growth with FLuid Attenuated Inversion Recovery (FLAIR) images. Instead, [5] monitored low-grade glioma growth via a dedicated segmentation pipeline, again on FLAIR images. Last, [6] tried to distinguish radiation-induced pseudo progression from real tumor progression using 3D shape features and a support vector machine. If we extend change detection also to remote sensing imaging, several works based on DL have shown promising results in recent years [48].

Several works have explored the potential of NLP for classifying radiology reports [18]–[29]. However, only a few studies later investigated the application of their trained report classifier for a downstream imaging task [21], [49]–[51]. The authors in [49] showed that their report classifier could be used to triage head MRI scans and identify relevant abnormalities. Instead, authors in [50] used labels generated from reports of FDG-PET/CT to detect and estimate the location of abnormalities in whole-body scans. The work [51] described an NLP model that is used to generate weak image-level labels which are later integrated into a semi-supervised framework for mass detection in mammography images. Last, the authors in [21] trained an NLP classifier to create weak labels from pathology reports and later used these weak labels to train a DL model on colon Whole Slide Images. Similarly to [21], [49]–[51], we investigate whether the report classifier built in [18] can be useful to generate weak labels which are then used for the downstream

---
[1] Although strictly speaking there is no transfer of knowledge for this subgroup, we loosely include Mixed Training among the TL types.

imaging task of glioma change detection.

Regarding the automation of TL, most works have focused on measuring *transferability* between domains and tasks: for instance [52] define transferability as the difference in performance between models trained on source and target tasks, and use this information to improve several downstream tasks using logistic regression models. Similarly, [53] proposed a computational approach to discover transferability between 26 CV tasks, yielding optimal combinations of DL features for each target task. In terms of domain choice, the authors in [54] propose an information theoretic framework which permits to rank convolutional neural networks trained on different source domains and understand which is the most suitable for knowledge transfer. Alternatively, [55] proposed an adversarial multi-armed bandit that automatically decides which (if any) are the features of the source network that are useful for the target network. Our work differs from the above since the automation of our TL pipeline is focused on the type of TL (fine-tuning, feature extraction, or mixed training) rather than the selection of the most relevant source domain or task.

Finally, previous works have already explored the influence of model size on TL [38], [39], [44]. However, these works focused on the *transductive* TL scenario [32] where $D_s \neq D_t$, whereas we found no work that assessed the impact of model size for inductive TL ($D_s = D_t$, $Y_s \neq Y_t$).

## 2. Materials and Methods

### 2.1 In-house Dataset

We retrieved 2100 MR scans belonging to 183 retrospective patients with high-grade gliomas who were scanned between 2004 and 2019 at the Lausanne University Hospital (average number of scans per subject 5, standard deviation 4.5). At every session, a series of MR scans including structural, perfusion and functional imaging were performed. For simplicity, in this work we only focused on the T2-weighted (T2w) scans. The MR acquisition parameters for the cohort are provided in Table 1. Scans that were too close to surgery (within 4 weeks) were excluded since they contained exaggerated intensity changes and brain deformations around the resection cavity. We deem these changes irrelevant since they are not related to the tumor itself. In addition, we extracted the radiology reports associated with each session. These were written (or dictated) in French during routine clinical practice by a junior radiologist after exploring all sequences of interest. Then, a senior radiologist reviewed each case amending the report when necessary. The extracted reports have varying length ranging from 121 to 751 words (average 325, standard deviation 84). The protocol of this study was approved by the regional ethics committee; written informed consent was waived. We release an anonymized version of our in-house dataset on Zenodo (DOI: 10.5281/zenodo.7214605) under the permissive CC BY 4.0 license [56].

### 2.2 BraTS Dataset

To assess the generalization of our pipeline to an external dataset, we ran inference on the longitudinal subjects of the Multimodal Brain Tumor Segmentation (BraTS) 2015 multi-institutional dataset. We selected the 2015 edition because it is the only one that contains patients with multiple scans (i.e. longitudinal patients). Out of the 20 available longitudinal patients, we discarded 5 because they only contained two scans, namely the one before tumor resection and the one right after. For the remaining 15 subjects, we used 59 MR scans (average of ~ 4 scans per subject), again only focusing on T2w scans. From these 59 scans, we generated 51 difference maps (creation process described in section 2.4) which were tagged by one radiologist with over 18 years of experience (P.H.), using the labels presented in section 2.3. We openly release these labels for other researchers (https://github.com/connectomicslab/Glioma_Change_Detection_T2w/blob/master/extra_files/df_dates_and_t2_labels_brats_tcia_2015.csv).

**Table 1**: MR acquisition parameters of the 2100 T2w scans belonging to the in-house patients used for the study

| # scans | Vendor | Model | Field Strength [$T$] | Median TR [$ms$] | Median TE [$ms$] | Median Voxel Spacing [$mm^3$] |
|---|---|---|---|---|---|---|
| 800 | Philips | Intera | 3.0 | 3000 | 80 | 0.45x0.45x4.0 |
| 445 | Siemens Healthineers | Skyra | 3.0 | 5000 | 77 | 0.45x0.45x3.3 |
| 304 | Siemens Healthineers | TrioTim | 3.0 | 4700 | 84 | 0.45x0.45x3.9 |
| 254 | Siemens Healthineers | Symphony | 1.5 | 4370 | 103 | 0.45x0.45x6.5 |
| 127 | Siemens Healthineers | Verio | 3.0 | 5000 | 85 | 0.45x0.45x3.3 |
| 85 | Siemens Healthineers | Aera | 1.5 | 6220 | 84 | 0.6x0.6x3.3 |
| 77 | Siemens Healthineers | Prisma | 3.0 | 4881 | 77 | 0.45x0.45x3.3 |
| 2 | Siemens Healthineers | Espree | 1.5 | 6000 | 93 | 0.53x0.53x7.2 |
| 2 | Philips | Ingenia | 1.5 | 3448 | 80 | 0.34x0.34x3.6 |
| 1 | Siemens Healthineers | Vida | 3.0 | 5320 | 77 | 0.45x0.45x3.3 |
| 1 | Philips | Achieva | 1.5 | 3659 | 50 | 0.39x0.39x4.55 |
| 1 | Philips | Panorama HFO | 1.0 | 5300 | 100 | 0.33x0.33x5.3 |
| 1 | GE HealthCare | Discovery MR750 | 3.0 | 7955 | 100 | 0.47x0.47x4.0 |

## 2.3 Report Tagging

From the 183 glioma patients of the in-house dataset, we created two sub-datasets: a Human-Annotated Dataset (HAD) and a Weakly-Annotated Dataset (WAD).

**Human Annotated Dataset (HAD)** - For this sub-dataset, three radiologists tagged the MR radiology reports with labels of interest. For each report, the annotators were instructed to assign 3 classes: one class that indicated the global conclusion of the report (global conclusion), one to indicate the evolution of the enhanced part of the tumor (T1w conclusion) and the last class to indicate the evolution of the tumor on T2-weighted sequences (T2w conclusion). For each of these classes, the annotator could choose between the following labels:

- **stable**: assigned when the tumor did not change significantly with respect to the previous comparative exam
- **progression**: assigned when the tumor worsened with respect to the previous comparative exam. This class included cases where the enhanced part of the tumor increased in size or when the T2 signal anomalies surrounding the tumor increased in extension
- **response**: assigned when the tumor responded positively to the treatment
- **unknown**: assigned if the annotator was not able to assign any of the three classes above when reading the report

381 reports (belonging to 169 distinct patients) were manually annotated by our three experts. Out of these 381, 39 reports (belonging to 39 distinct patients) were tagged by a senior radiologist with over 18 years of experience in neuroimaging (P.H), while 342 reports (belonging to 162 patients) were tagged by two radiologists both with 4 years of experience (C.A, E.G.T). Cohen's kappa coefficient between the two readers for the T2w conclusion was k=0.80 which is considered a "substantial agreement" [57]. The 41/342 reports for which the two annotators disagreed were discarded. Also, we discarded 90 reports for which the T2w conclusion was different from the global conclusion. Last, we also excluded the 17 reports for which the T2w conclusion was tagged as unknown. This left a total of 91 patients, 378 scans, and 233 difference maps (see Figure 1 for more details).

**Weakly Annotated Dataset (WAD)** - For this sub-dataset, reports were tagged with the classifier proposed in our previous work [18]. Briefly, this consists of an NLP pipeline in which we preprocess (e.g. removed proper nouns, stopwords, punctuation), embed (with TF-IDF) and then classify (with a Random

Forest) the radiology reports precisely into the classes mentioned above (i.e. **stable**, **progression** and **response**). We denote the labels generated from the report classifier as *weak* because the classifier will commit errors, and because, differently from human readers, it cannot abstain when the reports are unclear (i.e. there is no **unknown** label).

Both for **HAD** and **WAD** we merged **progression** and **response** into one unique class which we denote as **unstable**. By doing this, we narrowed the task to a binary classification problem where we try to distinguish between **stable** and **unstable** reports. After these modifications, HAD contained 233 reports (159 stable, 74 unstable), whereas WAD contained either 795 (333 stable, 462 unstable) or 361 (165 stable, 196 unstable) reports, depending on the probabilistic output of the random forest (hyperparameter *fraction_of_WAD*, details in section 2.6). A detailed overview of the dataset is provided in Figure 1.

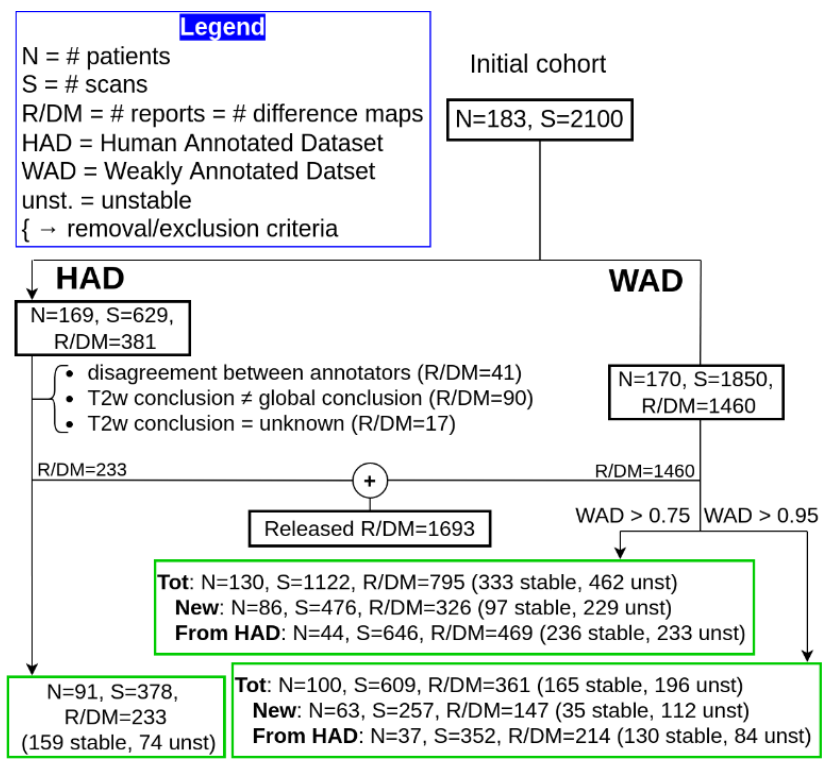

**Fig. 1**. (**COLOR**) Dataset overview. Each report corresponds to one T2w difference map since they both link two time points (i.e. two MR scans). The branches WAD > 0.75 and WAD > 0.95 depend on the hyperparameter *fraction_of_WAD* described in section 2.6. Difference maps in WAD come both from new patients (distinct from HAD patients) and HAD patients since not all reports from HAD have been tagged manually. Green rectangles indicate the final sets of patients/scans/difference maps used for the downstream analyses.

## 2.4 Image-Based Change Detection

While our former study [18] focused on report-based glioma change detection, this work deals with image-based glioma change detection. We know that every radiology report links two time points, namely the current scan and a previous scan which is used as baseline for comparison and longitudinal monitoring. Thus, for each report, we generated a corresponding T2w absolute difference map as illustrated in Figure 2. The rationale behind these difference maps is that parts of the tumor that either progress or respond to treatment (unstable) should appear as hyper-intense (examples (a) and (c) in Figure 2); instead, if the tumor is stable across the two time-points, the difference map will likely be hypo-intense overall (examples (b) and (d)). To generate the difference maps, we first applied N4 bias field correction with ANTs [58] both to the previous and to the current T2w volumes. Second, we registered the previous scan to the current scan, again with ANTs. Third, we skull-stripped both volumes (previous warped and current) with HD-BET [59]. Fourth, we applied z-score normalization on both volumes. Last, we computed the absolute voxel-wise

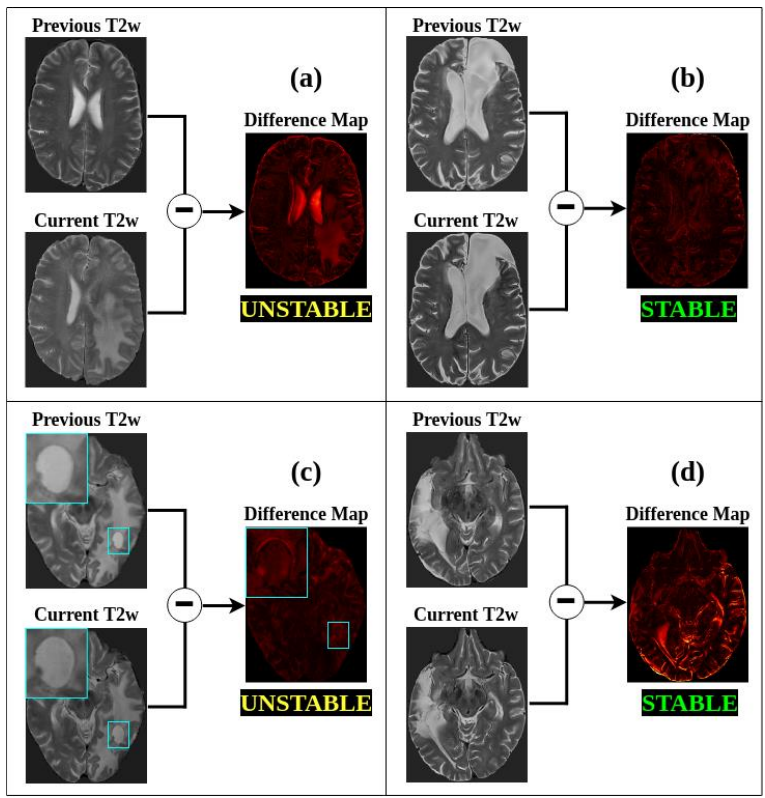

**Fig. 2**. (**COLOR**) Creation of T2w difference maps. After registration and normalization of the previous and current T2w volumes, the maps are computed via voxel-wise absolute difference. (a) 58-year-old male patient with a progressing (i.e. unstable) gliosarcoma. (b) 59-year-old male patient with a stable astrocytoma. (c) 60-year-old male patient with seemingly stable glioblastoma, but with enlarging cystic lesion (zoom inset in cyan color). (d) 60-year-old male patient with a (less evident) stable oligodendroglioma.

difference of the normalized volumes.

## 2.5 Classification Networks

The image-based change detection is treated as a binary classification problem: as for the reports, we try to classify the difference maps into stable and unstable. We used two Convolutional Neural Networks (CNNs) for the classification of the T2w difference maps: a custom 3D-VGG [60] (henceforth called VGG) and a 3D-ResNeXt [61] with Squeeze-and-Excitation [62] (henceforth called SEResNeXt). The VGG was written in PyTorch and contains 4 convolutional blocks followed by 4 fully-connected blocks. We used the ReLU activation function for all layers, except for the last layer which is followed by a sigmoid function. Batch normalization [63] was added in the VGG to prevent overfitting. The SEResNeXt was implemented with the MONAI framework [64]. For both CNNs, we used the cross-entropy loss function and the ADAM optimizer [65]. During training, we applied online data augmentations, namely flip, addition of Gaussian noise, zoom (from 0.7 to 1.3, 1 being the original volume size) and elastic deformation, each with probability of 20%. The total number of trainable parameters in our networks is ~7.5 M for the VGG and ~19.4 M for SEResNeXt. Training and evaluation were performed with PyTorch 1.11.0 and a GeForce RTX 3090 GPU.

## 2.6 Experiments and Hyperparameter Tuning

**Creation of weak labels with report classifier -** For this work, we adapted the report classifier [18] and trained it to classify the T2w conclusion (in [18] it was trained to classify the global conclusion). We ran a nested 5-fold cross-validation on the 233 HAD reports, selected the best hyperparameters, and finally performed inference with the best model on all WAD reports to obtain the weak labels later used for the image-based change detection.

**Image-based glioma change detection -** Because of computational constraints, we decided to fix some hyperparameters, and tune others. Among the fixed (not tuned) hyperparameters we chose a batch size of 4, and 60 training epochs with early stopping. Depending on the experiments detailed below, other hyperparameters were tuned using the Optuna framework [66] with default arguments (Tree-structured Parzen Estimator as sampler, and Median pruner), and maximizing the Area Under the Receiver Operating Characteristic Curve (AUC) of a dedicated validation set composed of 25% of the training subjects (details in section 2.7).

To understand which TL type is the most appropriate to improve classification performances and how model capacity can influence TL results, we performed two experiments (called Baseline and TL) with the two DL models described above (VGG and SEResNeXt): in the Baseline experiment, we conducted a 5-fold cross-validation only on HAD (details in section 2.7), and WAD was intentionally not used (i.e. no TL). Evaluation was performed on the test subjects of each cross-validation fold and then results were aggregated. The only two hyperparameters that were tuned for the Baseline experiment were *learning_rate* and *weight_decay*. The former was chosen from $\{10^{-4}, 10^{-5}, 10^{-6}\}$, whereas the latter was chosen from $\{0, 0.01\}$. Since only two hyperparameters were tuned, both for the VGG-Baseline experiment and the SEResNeXt-Baseline experiment we ran all the six hyperparameter combinations.

In the Transfer Learning (TL) experiment, we still performed a 5-fold cross-validation on HAD, but this time we also exploited the WAD difference maps. In addition to *learning_rate* and *weight_decay* (which were tuned identically to the Baseline), here we also searched for the best transfer learning configuration. Specifically, we tuned 3 additional hyperparameters: *mixed_training*, *feature_extraction* and *fraction_of_WAD*.

- *mixed_training* can either be True or False: if True, we use for training a mixed shuffled dataset that is composed of WAD difference maps and the difference maps of the training HAD patients (scenario 3, section 1); if instead *mixed_training* is False, we either:
    o Perform feature extraction if *feature_extraction* is True (scenario 2, section 1). This implies fine-tuning only the last fully connected layers of our CNNs.
    o Perform fine-tuning if *feature_extraction* is False (scenario 1, section 1). This implies fine-tuning all the layers of the CNNs.
- *fraction_of_WAD* indicates which portion of WAD to use. We added this hyperparameter because not all weakly-labeled data is necessarily useful. In other words, by tuning *fraction_of_WAD* we wanted to understand whether some reports (and hence some difference maps) are more informative than others. The tunable values that we chose for *fraction_of_WAD* were $\{WAD > 0.75, WAD > 0.95\}$ where 0.75 and 0.95 are the output probabilities (soft labels) of the report classifier from [18]. For instance, when using $WAD > 0.95$ we only use a small portion of WAD, namely only the reports for which the report classifier is highly confident (output probability $> 0.95$). Instead, when using $WAD > 0.75$ we also include reports for which the NLP classifier is less confident[2].

Figure 3 illustrates one branch of the tree containing all possible hyperparameter combinations for the TL experiment. Since running all combinations would have been computationally impractical, we only ran each TL experiment (VGG-TL and SEResNeXt-TL) for 4 days.

---

[2] In the beginning, we tried using all WAD, but this consistently led to lower performances (results not shown).

To summarize, we ran 4 experiments: VGG-Baseline, VGG-TL, SEResNeXt-Baseline, and SEResNeXt-TL. The comparisons between Baseline and TL aimed to assess the effectiveness of the weak labels in the WAD for improving classification results. Results related to the impact of weak labels and TL are reported in section 3.1. Instead, comparisons between the two CNNs (e.g. VGG-Baseline vs. SEResNeXt-Baseline) aimed to understand the influence that model size/capacity can have on TL strategies for our task. Results related to the impact of model size are reported in section 3.2. The most frequent hyperparameter combinations are then reported in section 3.3, and finally in section 3.4 we report inference results of our 4 models (VGG-Baseline, VGG-TL, SEResNeXt-Baseline, SEResNeXt-TL) on the longitudinal patients of the external BraTS 2015 dataset. For each model, we ran inference with the five trained model of the cross-validation, and then performed majority voting of the five predictions.

## 2.7 Cross-Validation and Evaluation

**Cross-Validation -** For the Baseline experiments, we performed a 5-fold cross-validation on HAD. At each cross-validation split, 80% (72/91 subjects, 166 difference maps) of the subjects were used to train the CNN (either VGG or SEResNeXt), while the remaining 20% (19/91 subjects; 67 difference maps) of the

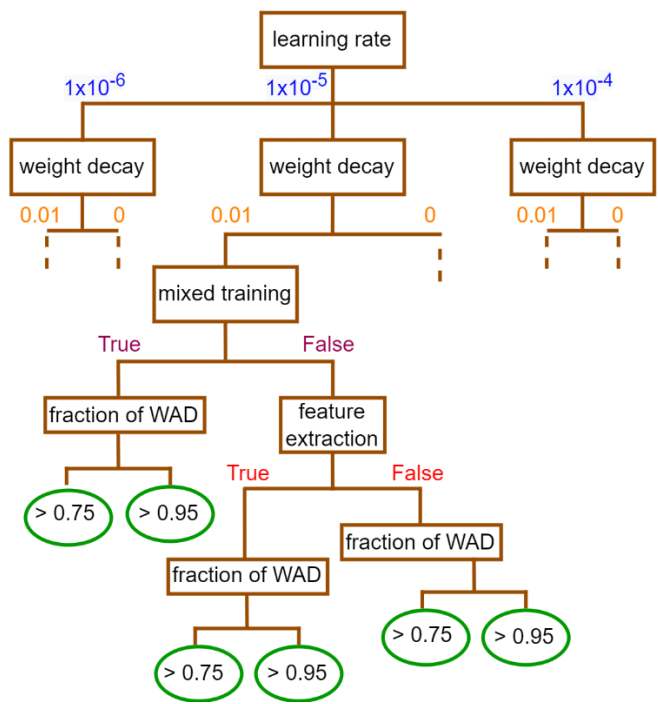

**Fig. 3**. (**COLOR**) Tree of hyperparameter combinations for the transfer learning experiments. Dashed lines indicate branches that are not shown because of limited figure space. Green ellipses are the leaves of the hyperparameter tree. If *mixed_training* is True, we end up in scenario 3 of section 1; if *mixed_training* is False and *feature_extraction* is True we perform feature extraction (scenario 2, section 1); if *mixed_training* is False and *feature_extraction* is False, we perform fine-tuning (scenario 1, section 1).

subjects were used to compute test results. Within each cross-validation fold, we also used 25% (18/72 subjects, 49 difference maps) of the training subjects as validation set for tuning the hyperparameters. To avoid over-optimistic results, the cross-validation splits were always performed at the subject-level to prevent multiple difference maps of the same subjects being assigned some to the training and some to the test (or validation) set. For the TL experiments, we performed the same 5-fold cross-validation on HAD, but then adapted the learning strategy according to the hyperparameters chosen during hypertuning: if *mixed_training* was True, then WAD subjects were added to the training subjects of HAD, while if *mixed_training* was False, the model was first pre-trained on WAD and then fully (fine-tuning) or partially

(feature extraction) fine-tuned on the training subjects of HAD. We ensured that the same splits were performed on HAD both for the Baseline and TL experiments, so that an exact comparison could be carried out. Also, for the HAD patients with overlapping scans (some manually annotated and some automatically annotated), we made sure to never assign some scans to training and some to test (or validation) set.

**Metrics -** The task that we address is binary classification of the T2w difference maps which are labeled either as stable or unstable. We report in the Results section accuracy, sensitivity (recall), specificity, F1 score, AUC, and Area Under the Precision-Recall curve (AUPR). We consider the class unstable as "positive", and the class stable as "negative".

**Statistics** - To statistically compare the four different models presented in section 3.6, we ran permutations tests using the difference in AUCs as test statistic, as similarly performed in [67]. We set a significance threshold $\alpha = 0.05$ and we ran 10,000 permutations for each test.

**Code availability -** All the code used for this paper is made available at https://github.com/connectomicslab/Glioma_Change_Detection_T2w, together with corresponding configuration files to reproduce the experiments.

# 3. Results

In cross-validation, the report classifier reached an accuracy of 93%, a sensitivity of 91% and a specificity of 94% on the 233 HAD reports. When running inference on WAD, 795 reports were associated with a class probability > 0.75, while 361 were associated with a class probability > 0.95.

The upper part of Table 2 shows test classification results of the VGG and the SEResNeXt for the task of image-based glioma change detection on the in-house dataset. The **Baseline** experiments are those where only HAD is used, while in the **TL** experiments we also leverage WAD. To visually summarize classification results, we also report in Figures 4 and 5 the Receiver Operating Characteristic (ROC) and the Precision-Recall (PR) curves, respectively.

## 3.1 Weak labels and TL improve classification results for VGG

We found a significant difference in AUC between the models VGG-Baseline and VGG-TL (p=0.05). This finding indicates the superiority of the TL pipeline with respect to the Baseline, which is visually confirmed in Figures 4 and 5 where the VGG-TL consistently outperforms VGG-Baseline.
Conversely, the permutation test indicated that the SEResNeXt-Baseline and SEResNeXt-TL had no significant difference (p=0.18), even though SEResNeXt-TL showed higher AUC and AUPR, and the corresponding PR curve outperforms the one from SEResNeXt-Baseline for most operating points. Overall, the two experiments show that only the VGG model benefits significantly from TL with the weakly-labeled dataset WAD.

## 3.2 Model size is negligible for the task at hand

To assess the impact of model size, we compared the VGG-Baseline vs. the SEResNeXt-Baseline model and found no significant difference between the two (p=0.17). Then, we also compared the VGG-TL to the SEResNeXt-TL model and again we found no significant difference (p=0.39). These experiments indicate that, for the task at hand, model size does not influence classification results, even though the SEResNeXt has ~ 2.5X more trainable parameters than the VGG (19.4M vs. 7.5M) and is slower to train (e.g. 1 epoch of the Baseline experiment takes 120 seconds for SEResNeXt vs. 90 seconds for VGG). Overall, these results suggest that the VGG is preferable for the task at hand because it is simpler and more computationally efficient.

## 3.3 Most frequent hyperparameters

Here, we report the most frequent hyperparameters that were chosen by the Optuna optimizer across the 5 training folds. For the VGG-Baseline experiment, the most frequent *learning_rate* was $10^{-4}$ (3 folds out of 5) and the most recurrent *weight_decay* was 0.01, while for the SEResNeXt-Baseline the most frequent *learning_rate* was $10^{-5}$ (3/5 folds) and the most frequent *weight_decay* was 0. More interestingly, we found a peculiar pattern in the hyperparameters of the TL pipeline: both for the VGG-TL (5/5 folds) and for the SEResNeXt-TL (4/5 folds), the hyperparameter *mixed_training* was always True. This means that training from scratch with a mixed dataset (WAD + training HAD) consistently leads to higher performances with respect to either fine-tuning or feature extraction. Regarding the hyperparameter *fraction_of_WAD*, the most frequent value for the VGG-TL experiment was WAD > 0.95 (3/5 folds), whereas the most recurrent value for SEResNeXt-TL was WAD > 0.75 (4/5 folds).

### 3.4 Inference on BraTS

Out of the 51 difference maps that we extracted from BraTS 2015, 12/51 (23%) were tagged as stable, while 39/51 (76%) were tagged as unstable, by our senior radiologist. The lower part of Table 2 illustrates inference results of the trained models after majority voting among the five splits of the cross-validation. Although the SEResNeXt-Baseline model showed the highest AUC, it did not significantly outperform the SEResNeXt-TL (p=0.46), or the VGG-Baseline (p=0.39).

**Table 2**. Classification test results. **Upper part**: in-house dataset. **Lower part**: BraTS-2015 dataset. Bold values indicate the highest performances. N=number of difference maps; Baseline=pipeline where only HAD data is used. TL=Transfer Learning: pipeline where both HAD and WAD are used. ACC=Accuracy; SENS=sensitivity; SPEC=specificity; F1=F1 score; AUC=Area Under ROC Curve; AUPR=Area Under the Precision-Recall curve; PARAMS=trainable parameters

| Dataset | N | MODEL | ACC | SENS | SPEC | F1 | AUC | AUPR | # PARAMS |
|---|---|---|---|---|---|---|---|---|---|
| In-house | 233 | VGG-Baseline | 70 | 55 | 77 | 54 | .74 | .55 | 7.5M |
|  |  | VGG-TL | **79** | **80** | 79 | **71** | .82 | .72 |  |
|  |  | SEResNeXt-Baseline | 76 | 50 | **88** | 57 | .79 | .63 | 19.4M |
|  |  | SEResNeXt-TL | 77 | 78 | 76 | 68 | **.83** | **.73** |  |
| BraTS 2015 | 51 | VGG-Baseline (inference) | 75 | 82 | 50 | 83 | .66 | .90 | 7.5M |
|  |  | VGG-TL (inference) | 76 | 92 | 25 | 86 | .59 | .89 |  |
|  |  | SEResNeXt-Baseline (inference) | 73 | 69 | **83** | 79 | **.76** | **.93** | 19.4M |
|  |  | SEResNeXt-TL (inference) | **78** | **95** | 25 | **87** | .60 | .60 |  |

## 4. Discussion

This work investigated the effectiveness of inductive TL for the task of image-based glioma change detection. To this end, we compared an automated TL pipeline that leverages weakly annotated data with a Baseline that uses only human-annotated data. The experiments were run with two CNNs (VGG and SEResNeXt) to assess the impact that model size can have on classification performances and finally the pipeline was validated on the external BraTS dataset to assess model generalizability.

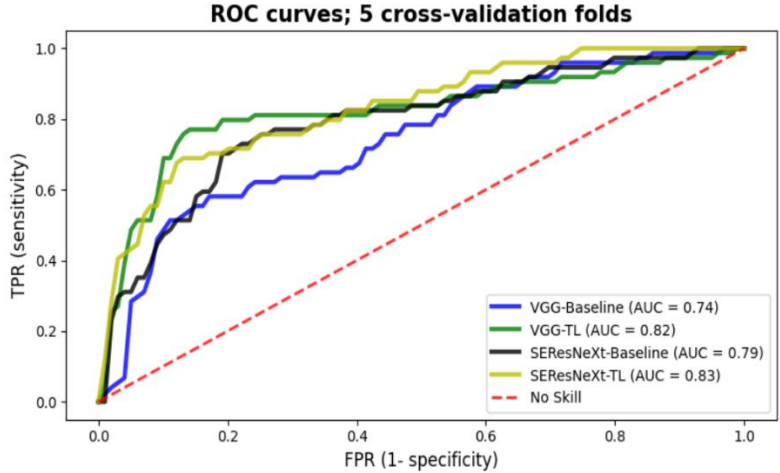

**Fig. 4** (**COLOR**). Receiver Operating Characteristic Curve (ROC) curves aggregated over the five test folds of HAD. AUC = Area Under the ROC Curve. TL = Transfer Learning.

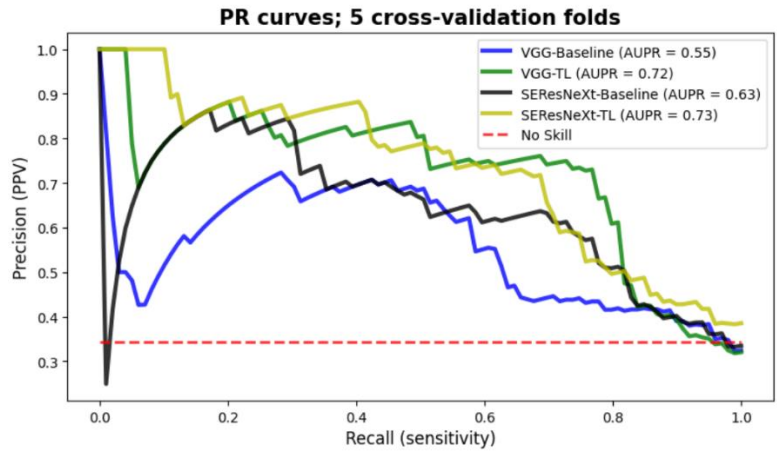

**Fig. 5** (**COLOR**). Precision-Recall (PR) curves aggregated over the five test folds of HAD. AUPR = Area Under the Precision Recall (PR) curve. TL = Transfer Learning.

Despite being less accurate, weak labels extracted from radiology reports hold great potential for mitigating the manual annotation bottleneck in medical imaging. The main advantage of NLP-generated weak labels is that report classifiers are normally fast to train (e.g. the one presented in [18] takes ~10 minutes), while DL-based image classifiers normally take several hours. Therefore, labeling hundreds (or even thousands) of new subjects becomes extremely fast and inexpensive. In this work, the weak annotation process allowed us to obtain a more than 3-fold increase in sample size (233 difference maps for HAD vs. 795 for the TL pipeline with $WAD > 0.75$) at very little added cost. Results in section 3.1 showed that the automatically-labeled dataset WAD helps improving classification results, although the difference in performance between Baseline and TL was only significant for the VGG model. This result differs from [38] since in the small data regime we found the smaller network (VGG) to benefit more from TL with respect to the larger SEResNeXt. Nonetheless, as similarly reported in [50], we expect performances of both models to increase even further as more weakly-labeled samples are added.

When studying the impact of model size in classification performances (section 3.2), we found no significant difference between VGG and SEResNeXt neither for the Baseline nor for the TL experiment. Therefore, for our application, we conclude that the VGG model is preferable because it is simpler and faster to train. A similar result was found in [68] where a VGG-19 model outperformed much deeper networks in a TL pipeline for COVID-19 detection. Both our results and the ones in [68] indicate that the high-capacity networks and transfer learning strategies typically used for computer vision tasks in the high-data regime are not necessarily optimal for medical imaging tasks, where models often operate in the low-data regime. Given that deep learning scaling studies typically show log-linear or power laws relating loss to dataset size [69], [70], it is possible that the higher-capacity SEResNeXt model in our study would be superior if much more data were available. However, this is not visible with our small dataset as we are far from the performance asymptote.

Another contribution of this work is the automation of the TL pipeline. Instead of searching for the best TL type manually, we framed the TL experiments as a hyperparameter optimization problem. We believe that our pipeline can be adopted by similar works that aim to automate TL for image classification. Surprisingly, we found that mixed training TL led to the highest classification performances. From a computational and environmental point of view, this finding is alarming because it indicates that the longest-running and least resource-efficient TL pipeline could be preferable with respect to feature extracting or fine-tuning.

As last contribution, we also evaluated our four models on the external BraTS dataset in order to assess model generalizability. Although the sample size is limited (51 difference maps) and no significant differences were found with the permutation tests, results on the lower part of Table 2 indicate that the two Baseline models (VGG and SEResNeXt) can better cope with class imbalance (higher specificity and AUC).

Our work has several limitations. First, we narrowed the classification problem to a binary scenario (**stable** vs. **unstable** tumor), mainly because we do not have enough cases of tumor **response** in our cohort. This is a simplification because **progression** and **response** are distinct clinical indicators. In future works, we are planning to extract new patients and adapt the classification towards a 3-class problem (**stable**, **progression**, **response**). As shown in [71], this will require careful analyses since results might change significantly when the labels of the task become more granular. The second limitation of this study is that we only focused on T2w MRI volumes, even though a multi-modal assessment of glioma evolution would be more accurate [10]. Third, the reports from the HAD for which the two annotators disagreed were discarded, while in the future we plan to use them after a consensus between the readers has been reached. Additionally, we only evaluated one approach for fine-tuning, whereas other strategies, including freezing different layers for different number of epochs [37], remain to be explored.

## 5. Conclusion

This study presented a TL pipeline that uses weakly-labeled data generated from radiology reports to improve classification performances for the task of glioma change detection. We found that a custom VGG model benefits more from transfer learning (and has similar performances) with respect to a more complex ResNet-like model. We hope this finding raises awareness regarding the potentially misleading translation between computer vision and medical imaging applications, and that our automated pipeline can be replicated for similar TL tasks in the field.

## Acknowledgment
We thank the Lundin Brain Tumor Research Center for support.